\documentclass[letterpaper]{article} 
\usepackage{aaai24}  
\usepackage{times}  
\usepackage{helvet}  
\usepackage{courier}  
\usepackage[hyphens]{url}  
\usepackage{graphicx} 
\usepackage{natbib}  
\usepackage{caption} 
\usepackage{algorithm}
\usepackage{algorithmic}
\usepackage{xcolor}
\usepackage{xspace}
\newcommand{\ourmethod}{PUCL\xspace}
\usepackage{amsmath}
\usepackage{amssymb}
\usepackage{mathtools}
\usepackage{amsthm}
\theoremstyle{plain}
\newtheorem{theorem}{Theorem}

\newtheorem{lemma}{Lemma}

\theoremstyle{definition}

\theoremstyle{remark}

\usepackage[capitalize,noabbrev]{cleveref}
\usepackage{microtype}
\usepackage{graphicx}
\usepackage{subfigure}
\usepackage{booktabs} 

\usepackage{algorithm}
\usepackage{algorithmic}

\usepackage{newfloat}
\usepackage{listings}
\usepackage{url}            
\usepackage{booktabs}       
\usepackage{amsfonts}       
\usepackage{nicefrac}       
\usepackage{microtype}      
\usepackage{xcolor}         

\usepackage{caption}
\usepackage{float}
\usepackage{times}

\usepackage{soul}
\usepackage{url}
\usepackage[utf8]{inputenc}
\usepackage{caption}
\usepackage{graphicx}
\usepackage{amsmath}
\usepackage{amssymb}
\usepackage{cuted}
\usepackage{booktabs}
\usepackage{newfloat}
\usepackage{listings}
\DeclareCaptionStyle{ruled}{labelfont=normalfont,labelsep=colon,strut=off} 
\floatstyle{ruled}
\newfloat{listing}{tb}{lst}{}
\floatname{listing}{Listing}
\title{Contrastive Learning with Negative Sampling Correction}
\author{Lu Wang$^*$, Chao Du$^*$, Pu Zhao$^*$,  Chuan Luo$^*$,   Zhangchi Zhu$^*$, Bo Qiao$^*$, Wei Zhang$^+$, Qingwei Lin$^*$,  Saravan Rajmohan$^\mathsection$, Dongmei Zhang$^*$, Qi Zhang$^*$}
\affiliations{
     Microsoft$^*$, Microsoft 365$^\mathsection$, East China Normal University$^+$
}

\usepackage{bibentry}

\begin{document}

\maketitle

\begin{abstract}
As one of the most effective self-supervised representation learning methods, contrastive learning (CL) relies on multiple negative pairs to contrast against each positive pair. In the standard practice of contrastive learning, data augmentation methods are utilized to generate both positive and negative pairs. While existing works have been focusing on improving the positive sampling, the negative sampling process is often overlooked. In fact, the generated negative samples are often polluted by positive samples, which leads to a biased loss and performance degradation. To correct the negative sampling bias, we propose a novel contrastive learning method named Positive-Unlabeled Contrastive Learning (PUCL). PUCL treats the generated negative samples as unlabeled samples and uses information from positive samples to correct bias in contrastive loss. We prove that the corrected loss used in PUCL only incurs a negligible bias compared to the unbiased contrastive loss. PUCL can be applied to general contrastive learning problems and outperforms state-of-the-art methods on various image and graph classification tasks. The code of \ourmethod is in the supplementary file.
\end{abstract}

\section{Introduction}
Self-supervised learning  with contrastive learning (CL) loss~\citep{oord2018representation,poole2018variational} has achieved remarkable success in representation learning for a variety of downstream tasks~\citep{chen2020simple,misra2020self}, such as image representation learning~\citep{misra2020self}, graph representation learning~\citep{sun2019infograph,you2020graph}, and language representation learning~\citep{logeswaran2018efficient}. 

Contrastive learning relies on the noise-contrastive estimation (NCE)~\cite{gutmann2010noise} in which  semantically similar (positive) and dissimilar (negative) pairs of data points are used for learning a representation that maps positive pairs close while pushing negative pairs apart in the embedding space \citep{chen2020simple,oord2018representation}. A standard CL method usually generates positive pairs $(x,x^+)$ via transformation on observation $x$ and sample negative pairs $(x,x^-)$ from the rest of training data (e.g., SimCLR~\citep{chen2020simple} and CMC~\citep{tian2020contrastive}). However, in real-world problems, the sampled ``negative'' pairs are often polluted by the positive ones. For example, in the illustrative example of image representation learning shown in Figure~\ref{fig:motivation}, two images of different dogs might be considered as negative pairs by the standard sampling method. Similarly, in graph representation learning, due to lack of exposure, two nodes with an unobserved link could be misinterpreted as negative pairs. This quality issue on negative sampling is a major concern for the performance of CL as contrast can be sensitive to the sample quality~\citep{chuang2020debiased,arora2019theoretical}. 




To improve the quality of the negative samples, Ring~\citep{wu2020conditional} samples negative ones conditionally in a ring around each positive one by predefined percentiles. However, it yields more bias than standard noise contrastive estimation. To reduce the bias of negative sampling, \cite{chuang2020debiased} and \cite{robinson2020contrastive} develop a debiased contrastive objective via estimating a negative distribution with the assumption that the unlabeled samples are sampled from the full data. Still, this assumption is generally untrue in practice, as the unlabeled data essentially form a biased subset of the full data without a significant portion of labeled positives, as shown in Figure~\ref{fig:motivation}.

In this paper, inspired by the Positive-Unlabeled Learning approach~\citep{chang2021positive,elkan2008learning}, we assume that only the positive and unlabeled samples are available for training and the distribution of unlabeled data differs from the full distribution. Under these  assumptions, we propose a new contrastive learning method named  Positive-Unlabeled Contrastive Learning  (\ourmethod) for correcting the negative sampling bias in the contrastive loss. This method only incurs negligible bias compared to the true unbiased loss in the supervised learning and is compatible with any algorithm that optimizes the standard contrastive loss.

We summarize our \textbf{contribution} as follows (1)
We develop a new contrastive learning method named \ourmethod in which the negative sample distribution is approximated based on the observed distribution of unlabeled and positive data. (2) We propose a contrast learning loss with correction for negative sampling bias and theoretically show that it only incurs negligible bias compared to the true unbiased loss in the supervised contrastive learning under reasonable conditions. (3) We empirically observe the performance improvement of \ourmethod over state-of-the-art approaches on image and graph classification.

\section{Related Work}
\subsection{Contrastive Representation Learning}
Contrastive learning (CL) ~\citep{oord2018representation} has been widely used in vision tasks~\citep{chen2020simple}, text~\citep{logeswaran2018efficient}, structural data like graphs~\citep{sun2019infograph}, reinforcement learning~\citep{srinivas2020curl}, and few-shot scenarios~\citep{khosla2020supervised}. A major challenge in CL lies in the generation of negative samples. SimCLR~\citep{chen2020simple} uses augmented views of other items in the same minibatch as negative samples. In \cite{wu2018unsupervised}, a memory bank that caches representations and a momentum update strategy are introduced to enable the use of an enormous number of negative samples. Ring~\citep{wu2020conditional} adopts a design in which negative ones are sampled conditionally in a ring around each positive by predefined percentiles. Apart from these designs, works like~\cite{wang2020understanding} studies the asymptotic properties of the contrastive scheme for optimizing the alignment of positive pairs and the uniformity of negative pairs. Still, those works did not address the fact that not all the pairs obtained in typical negative sampling methods are true negative pairs. To address the sampling bias, researchers start to take the viewpoint of negative sample correction to introduce correction on negative distribution  ~\citep{chang2021positive,chuang2020debiased,robinson2020contrastive}.


\subsection{Positive-Unlabeled Learning}

Focusing on tasks in which only the positive and unlabeled data are available for analysis, the study on Positive-Unlabeled learning has gained popularity in recent years. As summarized in  ~\cite{bekker2020learning}, existing methods on Positive-Unlabeled usually take one of the following two assumptions regarding how the positive ($P$) data and unlabelled ($U$) data are sampled: either both $P$ data and $U$ data are sampled from a single training set (\textbf{single-training-set} scenario), or they are independently drawn from a dataset with all positive samples and a dataset with all unlabeled samples (\textbf{case-control} scenario). As the case-control scenario implies that the unlabeled samples are sampled from the whole dataset, the single-training-set scenario in which the unlabeled samples are generated from a biased portion of the whole data is often considered as being more appropriate for typical contrastive learning tasks~\citep{chen2020simple,wu2018unsupervised}. 

Depending on how $U$ data is handled, existing Positive-Unlabeled methods can be divided into two categories. The first category is known as biased PU learning (e.g., \cite{liu2002partially}), in which a two-step method is utilized to (1) identify possible negative ($N$) data in $U$ data and (2) perform ordinary supervised learning on the $P$ data and $U$ data. The second category is the unbiased PU learning (e.g., \cite{lee2003learning,liu2003building}), in which $U$ data are down-weighted or corrected using the information on $P$ data to approximate the negative distribution. As a well-designed unbiased PU learning method can avoid the labeled noise issue commonly funded in biased PU learning methods, in this paper, we consider the unbiased PU learning method for estimating the negative distribution in contrastive learning tasks.

\section{Preliminary}
\label{Prelimilary}
\subsection{Modeling Positive and Unlabeled Data}
In a typical Positive-Unlabeled learning scenario, for a sample $x$, we may use indicator variable $y$ to represent whether $x$ is positive or not and indicator variable $s$ to represent whether $x$ is labeled. An unlabeled sample $x$ in PU learning has $s=0$ and unknown $y$, while a positive sample is always labeled with $s=1$ and $y=1$. To models the distributions of both positive and unlabeled samples, either the \emph{case-control scenario} or the \emph{single-training-set scenario}~\citep{bekker2020learning} can be used. 

The \emph{case-control scenario} assumes that the positive data are sampled from the positive distribution and the unlabeled data are independently sampled from the true population $p(x)$:
\begin{align}
    x|_{s=0}\sim p^u(x)\nonumber\sim  p(x) \sim \alpha  p^+(x) + (1-\alpha) p^-(x)\nonumber
\end{align}
where $\alpha=Pr(y=1)$ represents the prior of the positive samples in the true population, $p^+(x)$ and $p^-(x)$ represent the distributions of positives and negatives in the full population respectively.  


The \emph{single-training-set scenario} assumes an initial dataset is firstly generated as i.i.d. samples from the true distribution. A fraction of the positive samples in this dataset are then selected and labeled as  the positive data. The unlabeled data is then sampled from the rest unlabeled samples. Under this scenario, the distribution of the initial dataset is identical to the population distribution $p(x)$, which can be represented either as a mixture of the positive and negative samples or as a mixture of positive labeled samples with distribution $p^l(x)$ and the unlabeled samples with distribution $p^u(x)$:
\begin{align}
    x\sim p(x)\sim \alpha p^+(x) + (1-\alpha)p^-(x) \sim \alpha c p^l(x) \nonumber\\  + (1-\alpha c) p^u(x),
    \label{distribution:unlabeled}
\end{align}
where $c = Pr(s=1|y=1)$ represents the fraction of positive cases to be labeled, and $\alpha c$ thus indicates the fraction of labeled positive samples. 

A key difference between these two scenarios is that the population of unlabeled data in the case-control scenario is identical to the true population ($p^u\sim p$) while the population of unlabeled data in the single-training set scenario only constitutes a biased portion of the true population ($p^u\neq p$).

\subsection{Contrastive Learning in Positive-Unlabeled Learning Framework}

\begin{figure}[h]
  \begin{center}
    \includegraphics[width=0.5\textwidth]{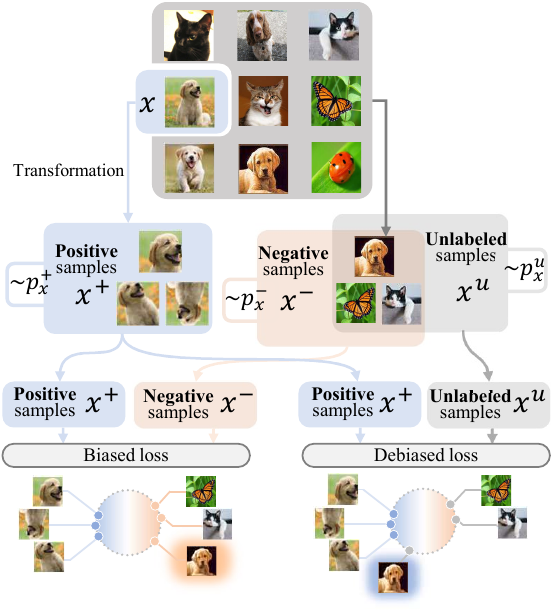}
  \end{center}
    \caption{ An illustration on the generation of positive and negative/unlabeled data in common contrastive learning. Treating unlabeled data as negatives can lead to biased loss when some positive samples are mislabeled (the dog on the right). While unlabeled data $x^u$ are sampled from partial data of $p$, a debiased loss can be formed with the correct treatment of the unlabeled samples.}
  \label{fig:motivation}
\end{figure}

In the unsupervised setting of standard contrastive learning, for an observation $x$ from set $\mathcal{X}$ , new samples can be generated by applying semantic-preserving transformations on $x$ such as random cropping, color jitter, Gaussian blurring, etc. in computer vision. We will use $\mathcal{T}(x,\epsilon)$ to denote a semantics-preserving transformations of $x$ with random element $\epsilon \sim p(\epsilon)$. For each observation $x$ sampled from the data distribution $p(x)$, a positive pair $(x,x^+)$ with semantically similar samples can be generated by sampling $x^+$ from the associated positive distribution $p_x^+$ via the semantics-preserving transformations $x^+=\mathcal{T}(x,\epsilon)$. While a negative pair $(x,x^-)$ of semantically dissimilar samples can be in principle generated by sampling $x^-$ from the negative distribution $p_x^-$, $p_x^-$ is typically not accessible in practice. Standard contrastive learning method usually generates a ``negative'' pair $(x,x^-)$ by sampling $x^-$ from set $\mathcal{X}\setminus\{x, x^+=\mathcal{T}(x,\epsilon)\}$ that excludes $x^+$ generated via the semantics-preserving transformations. Still,  $\mathcal{X}\setminus\{x, x^+=\mathcal{T}(x,\epsilon)\}$ could include both samples semantically dissimilar and semantically similar to $x$. From the view point of Positive-Unlabeled learning, a ``negative'' pair $(x,x^-)$ in standard contrastive learning should be regarded as an unlabeled pair $(x,x^u)$ with $x^u$ sampled from the unlabeled distribution $p_x^u$. This process of generating both positive and unlabeled (``negative'' as in standard contrastive learning) data is illustrated in Figure \ref{fig:motivation}. 

Therefore, in contrast to the standard contrastive learning that requires samples from both positive and unlabeled distributions, our reformulation of contrastive learning in the Positive-Unlabeled learning framework assumes that we can only generate samples from the positive data consisting of semantically similar pairs $(x,x^+)$ and the unlabeled data consisting of pairs $(x,x^u)$. And our goal of the study is still to learn an encoder $f$ capable of differentiating the semantically similar pairs from the semantically dissimilar pairs for the down streaming task, such as classification. 


\textbf{The single-training-set scenario is more appropriate for the contrastive learning.} Similar to the single-training-set scenario, the unlabeled data $(x,x^u)$ in contrastive learning is sampled from a biased portion of the full dataset that excludes positive pairs generated via applying semantics-preserving transformations on $x$. Moreover, instead of sampling positive and unlabeled data independently from two populations as in the case-control scenario, the sampling of positive pairs is usually strongly correlated with the sampling of unlabeled pairs in a typical contrastive learning setting. Therefore, we believe that the single-training-set scenario is more appropriate for contrastive learning and will adopt this assumption to model the distributions of positive and unlabeled data throughout this paper.

\subsection{Negative Sampling Bias in Contrastive Loss}

In standard contrastive learning, for each observation $x$, one positive pair $(x,x^+)$ and $N$ independent negative pairs $\{(x,x_i^-)\}_{1:N}$ can be generated. Then given an encoder $f: \mathbb{R}^n \xrightarrow{}\mathcal{S}^{d-1}$ that maps each observation to a d-dimensional feature vector with unit norm, we define the contrastive loss as: 
{\small
\begin{align}
\mathcal{L}_{\textit{IdealCL}} = \mathbb{E}_{\begin{subarray}{l}x\sim p, x^+\sim p^+_x,
\{x_i^-\}_{i=1}^N\sim p^{-N}_x\end{subarray}}
[-\log \frac{h(x,x^+)}{h(x,x^+)+V^-}],\nonumber
\end{align}
}
\noindent where $h(x,x_i)=e^{f(x)^\mathsf{T}f(x_i)}$, $V^- = \sum_{i=1}^N h(x,x^-_i)$ and $\{x_i^-\}_{i=1}^N\sim p^{-N}_x$ indicates that the $N$ negative pairs are sampled from negative distribution $p^{-}_x$ independently. 


However, as discussed in the previous section, without direct access to $p^-_x$, the standard contrastive learning method can only obtain samples from the unlabeled distribution $p^u_x$. By treating $\{x^u_i\}_1^N$ as $\{x^-_i\}_1^N$, the loss used in standard contrastive learning is essentially a biased version of the ideal loss $\mathcal{L}_{\textit{IdealCL}}$. We will term this biased loss as the standard contrastive loss $\mathcal{L}_{staCL}$: 
{\small
\begin{align}
 \mathcal{L}_{staCL} \nonumber = \mathbb{E}_{\begin{smallmatrix}x\sim p, x^+\sim p^+_x, \{x_i^u\}_{i=1}^N\sim p^{uN}_x\end{smallmatrix}}[-\log \frac{h(x,x^+)}{h(x,x^+)+V^u}],   
 \end{align}
}
where $V^u = \sum_{i=1}^N h(x,x^u_i)$ and $\{x_i^u\}_{i=1}^N$ are i.i.d. samples from the unlabeled distribution. Compared to the unbiased ideal loss $\mathcal{L}_{\textit{IdealCL}}$, the use of $\mathcal{L}_{staCL}$ would introduce non-negligible negative sampling bias which reduces the effectiveness of contrastive learning method~\cite{chuang2020debiased}.

\section{Method}\label{method}


In this section, we analyze an alternative loss to the ideal loss $\mathcal{L}_{\textit{IdealCL}}$ with correction for negative sampling bias. We show that, this loss forms a debiased estimation of the ideal loss and can be efficiently estimated using only positive and unlabeled samples in the setting of contrastive learning.




\subsection{Representation of Negative Distribution}
To circumvent the unavailability of negative distribution in estimating the contrastive loss, we note that the theoretical framework of Positive-Unlabeled learning can be leveraged so that the negative distribution can be estimated via the available positive and unlabeled distributions. As discussed in Section \ref{Prelimilary}, we will adopt the single-training-set scenario specified by Equation~\ref{distribution:unlabeled} to model the relationship between positive, negative, and unlabeled distributions. A direct transformation of Equation~\ref{distribution:unlabeled} allows us to represent the negative distribution as follows:


{\small
\begin{align}
   p^-_x(x') = \frac{1-\alpha c}{1-\alpha}p^u_x(x') - \frac{\alpha}{1-\alpha}p^+_x(x')+\frac{\alpha c}{1-\alpha}p^l_x(x'),
    \label{unlabeld theorem}
\end{align}
}

This representation of the negative distribution can be further simplified by additional assumptions on the labeling mechanism. In this paper, we will follow the convention to assume that the labeling mechanism satisfies the Selected Completely At Random (SCAR) assumption.

\textbf{Definition 1 (Selected Completely At Random (SCAR))} A labeling mechanism is SCAR if all labeled samples are selected completely at random from the positive distribution independent from their attributes. Under SCAR, the propensity score $e(x)$, the probability for labeling a positive sample, is constant and equal to $c$, the proportion of labeled samples among positive samples: 
\begin{align}
e(x) = Pr(s = 1|x, y = 1) = Pr(s = 1|y = 1) = c.\nonumber
\end{align}
Under this assumption, labeled data can be viewed as i.i.d. samples from the positive distribution with $p^l_x = p^+_x$. Equation \ref{unlabeld theorem} can then be further simplified into the following lemma: 
\begin{lemma} Under the single-training-set scenario and the SCAR assumption, the negative sample distribution can be represented by the unlabeled and positive distributions as following:
\begin{align}
   p^-_x(x^-) = \frac{1-\alpha c}{1-\alpha}p^u_x(x^-) - \frac{\alpha(1-c)}{1-\alpha}p^+_x(x^-).
\end{align}
\label{the:neg}
\end{lemma}
Lemma~\ref{the:neg} clearly shows the distinction between the unlabeled and negative distributions in contrastive learning. It also suggests that the negative distribution can be estimated unbiased through correcting the unlabeled distribution with information from the positive distribution. 

\subsection{Correcting Contrastive Loss with Positive and Unlabeled Data}

By Lemma ~\ref{the:neg}, the joint distribution of $N$ negative samples $p^{-N}_x$ can be represented as $[\frac{1-\alpha c}{1-\alpha}p^u_x - \frac{\alpha(1-c)}{1-\alpha}p^+_x]^N$. While this representation can be leveraged to rewrite $\mathcal{L}_{\textit{IdealCL}}$ as an expectation over the unlabeled and positive distributions, it would also incur a high computational cost for evaluating expectation with respect to every  terms in the expansion of $[\frac{1-\alpha c}{1-\alpha}p^u_x - \frac{\alpha(1-c)}{1-\alpha}p^+_x]^N$. Here we construct an alternative of the contrastive loss $\mathcal{L}_{\textit{IdealCL}}$ named as $\mathcal{L}_{\textit{DeCL}}$. This new loss is obtained by replacing the term $\sum_{i=1}^N h(x,x^-_i)$ in $\mathcal{L}_{\textit{IdealCL}}$ with its expectation $N\mu_x$ where $\mu_x= \mathbb{E}_{x^-\sim p_x^-}[h(x,x^-)]$. The following theorem shows that the difference between $\mathcal{L}_{\textit{DeCL}}$ and $\mathcal{L}_{\textit{IdealCL}}$ is negligible for large $N$ and $\mathcal{L}_{\textit{DeCL}}$ can be computed direct from the positive and unlabeled distributions. 
\begin{theorem}
\label{theo:debias}
{\small
\begin{align}
 \label{obj:appcl}
     &|\mathcal{L}_{\textit{IdealCL}}-\mathcal{L}_{\textit{DeCL}}|\leq \frac{1}{2\sqrt{N}} (e^{2}-1),\\ & \text{where } \mathcal{L}_{\textit{DeCL}}= \mathbb{E}_{x\sim p, x^+\sim p^+_x}[-\log
     \frac{h(x,x^+)}{h(x,x^+)+N\mu_x}] \\ & \text{ and } \mu_x = \mathbb{E}_{x^-\sim p^-}[h(x,x^-)]=\frac{1-\alpha c}{1-\alpha}\mathbb{E}_{x^-\sim p^u}[h(x,x^-)] \\ & -\frac{\alpha(1-c)}{1-\alpha}\mathbb{E}_{x^+\sim p^+_x}[h(x,x^+)] \nonumber.
\end{align}
}
\end{theorem}

Theorem \ref{theo:debias} shows that $\mathcal{L}_{\textit{DeCL}}$ only incurs a negligible bias compared to the ideal loss $\mathcal{L}_{\textit{IdealCL}}$ as large $N$ is often used in standard contrastive learning methods. As $\mu_x$ can be represented as an expectation over the unlabeled distribution corrected by a term based on the positive distribution, $\mathcal{L}_{\textit{DeCL}}$ can be estimated using only positive and unlabeled samples. In practice, the following empirical estimate will be used for estimating $\mathcal{L}_{\textit{DeCL}}$:
{\small
\begin{align}
 \label{obj:appcl}
     &L_{\textit{DeCL}}^{M^u,M^+} = \mathbb{E}_{x\sim p, x^+\sim p^+_x}[-\log
     \frac{h(x,x^+)}{h(x,x^+)+N\hat{\mu}_x}],\\
     &\text{where }\hat{\mu}_x = \max\{ \frac{1-\alpha c}{1-\alpha}[\frac{1}{M^u}\sum_{i=1}^{M^u} h(x,x^u_i)] \\&- \frac{\alpha(1-c)}{1-\alpha}[\frac{1}{M^+}\sum_{j=1}^{M^+} h(x,x^+_j)], e^{-1} \}\nonumber,
\end{align}
}
where $\{x^u_i\}^{M^u}_{i=1}$ are $M^u$ i.i.d. samples from the unlabeled distributions $p^u_x$ and $\{x^+_j\}^{M^+}_{j=1}$ are $M^+$ i.i.d. samples from the positive distribution $p^+_x$. The maximum is taken to ensure that $\hat{\mu}_x$ is always no less than $e^{-1}$, the theoretical minimum of $\mu_x$. The proof of Theorem~\ref{theo:debias} can be found in Appendix B.

\section{Experiments}
In this section, in order to evaluate the effectiveness of \ourmethod, we first perform thorough experiments on image classification and graph classification tasks to compare \ourmethod against four state-of-the-art approaches. Then we conduct more empirical evaluations to analyze the effects of our \ourmethod.
\subsection{Experimental Settings}
\textbf{Dataset Description.}\label{dataset desc}
We evaluate \ourmethod empirically on image and graph representation tasks. For image representation, we conduct experiments on three datasets, including CIFAR10~\citep{krizhevsky2009learning} that consists of $60,000$ images with $10$ classes, CIFAR100~\citep{krizhevsky2009learning} which consists of $60,000$ images with 100 classes and STL10~\citep{coates2011analysis} that has $5000$ labeled images with $10$ classes and $100,000$ unlabeled images for unsupervised learning. 
For graph representation learning, we conduct experiments on four datasets, i.e., PTC, ENZYMES, DD, REDDIT-BINARY, from TUDataset~\citep{morris2020tudataset}. Detailed summary statistics of all datasets can be found in Appendix B.2. 
Following the previous work~\citep{robinson2020contrastive,wu2020conditional}, we use the \textbf{classification accuracy (Acc)} of fine-tuning a classifier on fixed embedding of data as the metric of both image task and graph task. All experimental details can be found in Appendix B. 

\begin{table}
    \centering
    \caption{Image classification accuracy (Acc) on CIFAR10, CIFAR100 and STL10.} 
    \label{image acc}
    \subtable[SimCLR]{
        \begin{tabular}{cccc}
        \toprule
        \texttt{Model}  & \texttt{CIFAR10} &\texttt{CIFAR100} &\texttt{STL10} \\
        \hline
        \texttt{SimCLR}  & 90.44    &64.94  &  80.50 \\
         \texttt{SimCLR+DEB}  & 90.21 &  66.55 & 82.17\\
        \texttt{SimCLR+Hard}  & 90.87 &  67.10 & 83.33\\
        \texttt{SimCLR+Ring} & 91.48 & 66.21 & 83.95 \\
        \texttt{SimCLR+\ourmethod} & \textbf{92.14} & \textbf{68.14} & \textbf{85.62}\\ 
        \bottomrule
        \end{tabular}
    }
    \subtable[CMC]{
        \begin{tabular}{cccc}
        \toprule
        \texttt{Model}  & \texttt{CIFAR10} &\texttt{CIFAR100} &\texttt{STL10} \\
        \hline
        \texttt{CMC}  & 87.20    &58.97  &  77.99 \\
         \texttt{CMC+DEB}  & 88.12 &  61.05 & 80.32\\
        \texttt{CMC+Hard}  & 88.29 &  62.04 & 80.68\\
        \texttt{CMC+Ring} & 88.14 & 60.99 & 80.62\\
        \texttt{CMC+\ourmethod} & \textbf{89.25} & \textbf{63.16} & \textbf{81.30}\\ 
        \bottomrule
        \end{tabular}
    }
    \subtable[MoCo]{
        \begin{tabular}{cccc}
        \toprule
        \texttt{Model}  & \texttt{CIFAR10} &\texttt{CIFAR100} &\texttt{STL10} \\
        \hline
        \texttt{MoCo}  & 85.62    &66.38  &  89.13 \\
        \texttt{MoCo+DEB}  & 88.36 &  67.31 & 91.87\\
        \texttt{MoCo+Hard}  & 89.99 &  68.02 & 92.71\\
        \texttt{MoCo+Ring} & 86.47 & 66.96 & 90.78\\
        \texttt{MoCo+\ourmethod} & \textbf{90.98} & \textbf{71.21} & \textbf{93.68}\\ 
        \bottomrule
        \end{tabular}
    }
\end{table}

\textbf{Baselines and Competitors.}\label{baseline}
To evaluate the negative sampling methods on image representation tasks, we use \textbf{SimCLR}~\citep{chen2020simple}, \textbf{CMC}~\citep{tian2020contrastive} and \textbf{MoCo}~\citep{he2020momentum} as the baseline models, which use a uniform distribution to sample the negative samples from unlabeled data.

As for the graph representation learning task, we use the state-of-the-art method \textbf{InfoGraph}~\citep{sun2019infograph} as the baseline, which is suitable for the downstream graph-level classification. 

We also compare \ourmethod with 2 state-of-the-art negative sample methods for contrastive learning: \textbf{Ring} ~\cite{wu2020conditional} samples negatives within a certain range to get rid of the negative pair that are either too far or too close; \textbf{Hard} ~\cite{robinson2020contrastive}, an improvement of \textbf{DEB} \cite{chuang2020debiased}, samples negatives from a self-designed negative distribution that gives more weights for hard negatives rather than sampling negatives uniformly from the training data.

For all the baselines, we carefully tuned the hyperparameters to achieve the best performances. The details of how we implement baselines and competitor can be found in Appendix B.

\subsection{Image Representation Learning}
First, for CIFAR10~\cite{krizhevsky2009learning}, CIFAR100~\cite{krizhevsky2009learning} and STL10~\cite{coates2011analysis}, we use SimCLR~\cite{chen2020simple}, CMC~\cite{tian2020contrastive}, and MoCo~\citep{he2020momentum} as baselines and compare \ourmethod with different SOTA negative sampling methods over those baselines. For each baseline, we set the size of negative samples $N$ differently depending on the designs of baseline models. Specifically, SimCLR uses a \textbf{batch-level negative sample} approach in which data in the same training batch are viewed as negative samples. Thus, in SimCLR, the negative sample size $N$ is related to the batch size and is set as 510. CMC uses a memory bank to save the whole training dataset and samples many more negative samples compared to SimCLR at each time step. We name this approach as  \textbf{memory-bank level negative sample} and set $N$ = 4096. MoCo encodes the keys with a momentum-updated encoder and maintains a queue of data samples. The number of negative samples is equivalent to the size of the queue. We name this approach as \textbf{queue-level negative sample} and set $N$ = 4096. In all those three baselines, for \ourmethod, we set the size of unlabeled samples $M^u=N$ for a fair comparison between \ourmethod and other SOTA negative sampling methods. The detailed experimental settings can be found in the Appendix A. 

\subsubsection{Results for Baseline SimCLR ($N$=$510$)}
Table~\ref{image acc} reports the Acc of the four methods on CIFAR10, CIFAR100 and STL10, using SimCLR as the baseline method. We use ResNet-50~\cite{he2016deep} as the encoder architecture. 
It shows \ourmethod achieves the state-of-the-art results with SimCLR. 

We observe that (1) \ourmethod consistently outperforms all competitors on all the image tasks, which verifies \ourmethod's effectiveness in estimating negative distribution. (2) Although different negative sampling competitors show different performances on these images, all of them outperform the baseline \emph{SimCLR}. The first phenomenon shows that estimating negatives plays an important role in contrastive learning and can significantly impact performance. The second phenomenon indicates that not all the unlabeled data are negatives and a well-designed negative sampling strategy can help choose the true negatives. 

(3) Compared with \emph{Hard}, which assumes the unlabeled data are sampled from the whole datasets, \ourmethod shows a significant improvement, which testifies the need of adopting a correct assumption on the unlabeled distribution. (4) The performance of \emph{Ring} is not stable among different image tasks. This might occur due to the fact that \emph{Ring} is sensitive to the percentile of the negative boundary, which is hard to tune. (5) We also observe that SimCLR can benefit from our method with improvements of 1.7\%, 3.2\%, and 5.12\% on CIFAR10, CIFAR100, and STL10, respectively, which verifies the effectiveness of \ourmethod. The improvement of \ourmethod over \emph{DEB}, \emph{Hard} and \emph{Ring} is also significant. In particular, on {STL10}, \ourmethod achieves 4.19\%, 2.48\% and 1.98\% improvements compared to \emph{Hard} and \emph{Ring} with SimCLR as the baseline model.%

\begin{table}[t]
    \caption{Graph classification accuracy (Acc) on PTC, ENZYMES, DD and REDDIT-BINARY.}
    \label{graph acc}
    \centering
        \subtable[PTC, ENZYMES]{
        \begin{tabular}{ccc}
        \toprule
        \texttt{Model}  & \texttt{PTC}& \texttt{ENZYMES}  \\
        \hline
        \texttt{InfoGraph}  & 56.63 & 50.33\\
        \texttt{InfoGraph+DEB}  & 55.74 & 51.21\\
        \texttt{InfoGraph+Hard}  & 56.20 & 51.35\\
        \texttt{InfoGraph+Ring} & 57.67 & 54.15 \\
        \texttt{InfoGraph+\ourmethod{}($\alpha$=$0.1$)}  & \textbf{58.90} & \textbf{56.21} \\
        \texttt{InfoGraph+\ourmethod{}($\alpha$=$0.5$)}  & 57.91 & 55.35 \\
        \bottomrule
        \end{tabular}
    }
    \subtable[DD, REDDIT-BINARY ]{
        \begin{tabular}{ccc}
        \toprule
        \texttt{Model} & \texttt{DD} & \texttt{REDDIT-B} \\
        \hline
        \texttt{InfoGraph }  & 73.16 & 82.99 \\
        \texttt{InfoGraph+DEB} & 70.58  & 84.89\\
        \texttt{InfoGraph+Hard} & 71.61  & 85.07\\
        \texttt{InfoGraph+Ring} & 76.32 & 85.30 \\
        \texttt{InfoGraph+\ourmethod{}($\alpha$=$0.1$)} & \textbf{77.62} & 87.08\\
        \texttt{InfoGraph+\ourmethod{}($\alpha$=$0.5$)} & 77.23 & \textbf{88.97}\\
        \bottomrule 
        \end{tabular}
    }
\end{table}

\begin{figure*}[!t]
\centering
\includegraphics[width=1\textwidth]{ 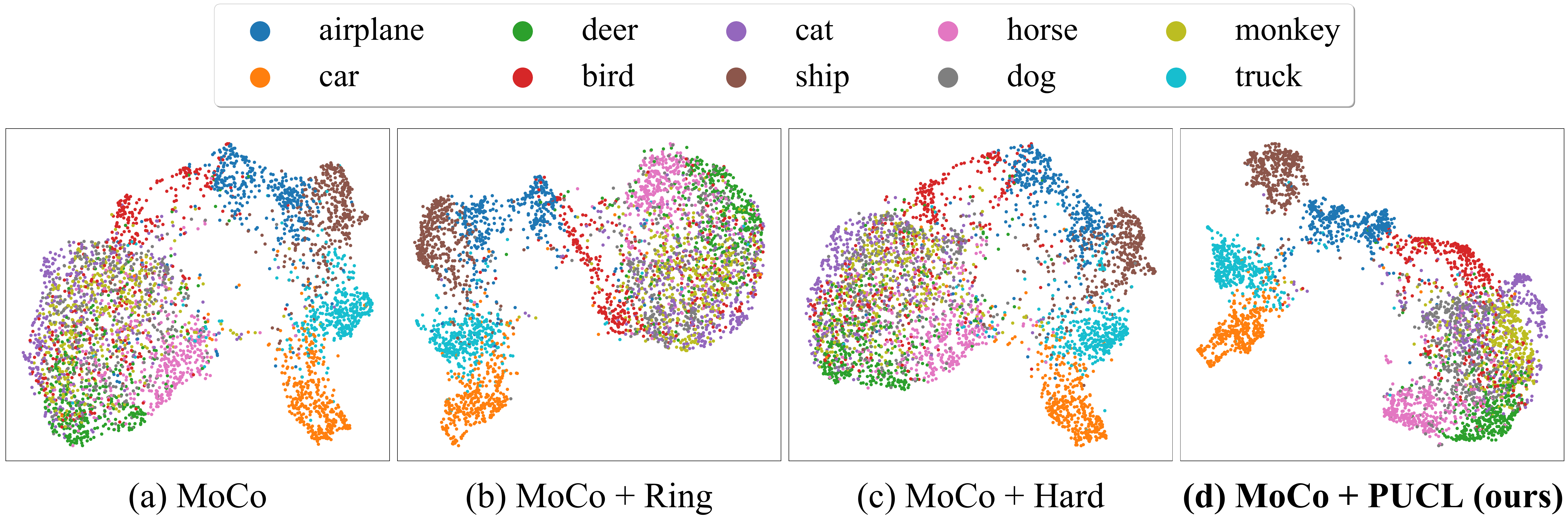}
\caption{Embedding visualization.}
\label{fig:emdvis}
\end{figure*}
\begin{figure} [th]
\centering
\hspace{-1.5em}
\subfigure[Different unlabeled sample size $M^u$ and $\alpha$]{
 \centering
 \includegraphics[width=0.48\textwidth, trim={1cm 0 1cm 0}]{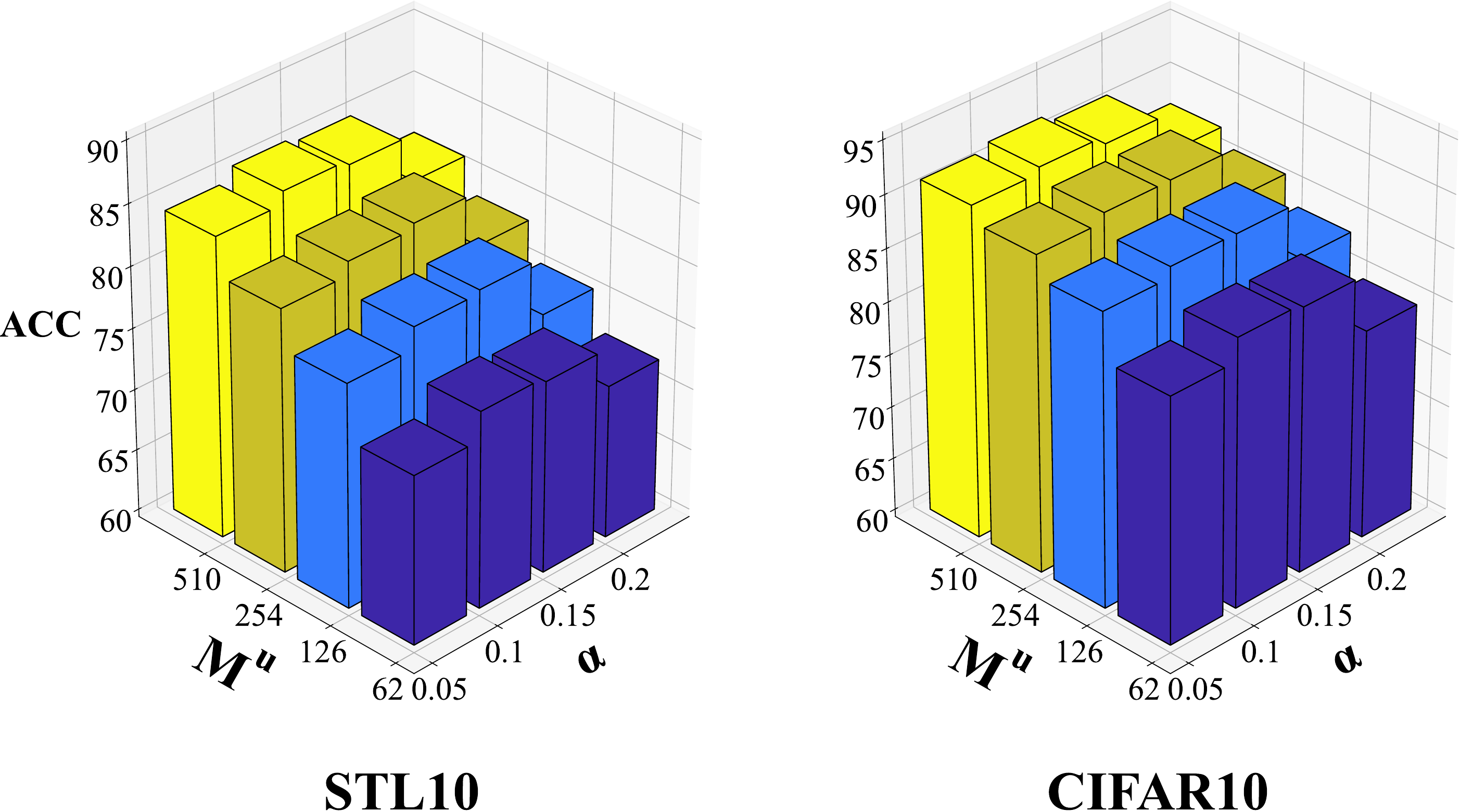}
 \label{fig:negab}
}
\subfigure[Different $\alpha$ and $c$]{
 \centering
 \includegraphics[width=0.48\textwidth, trim={1cm 0 1cm 0}]{ 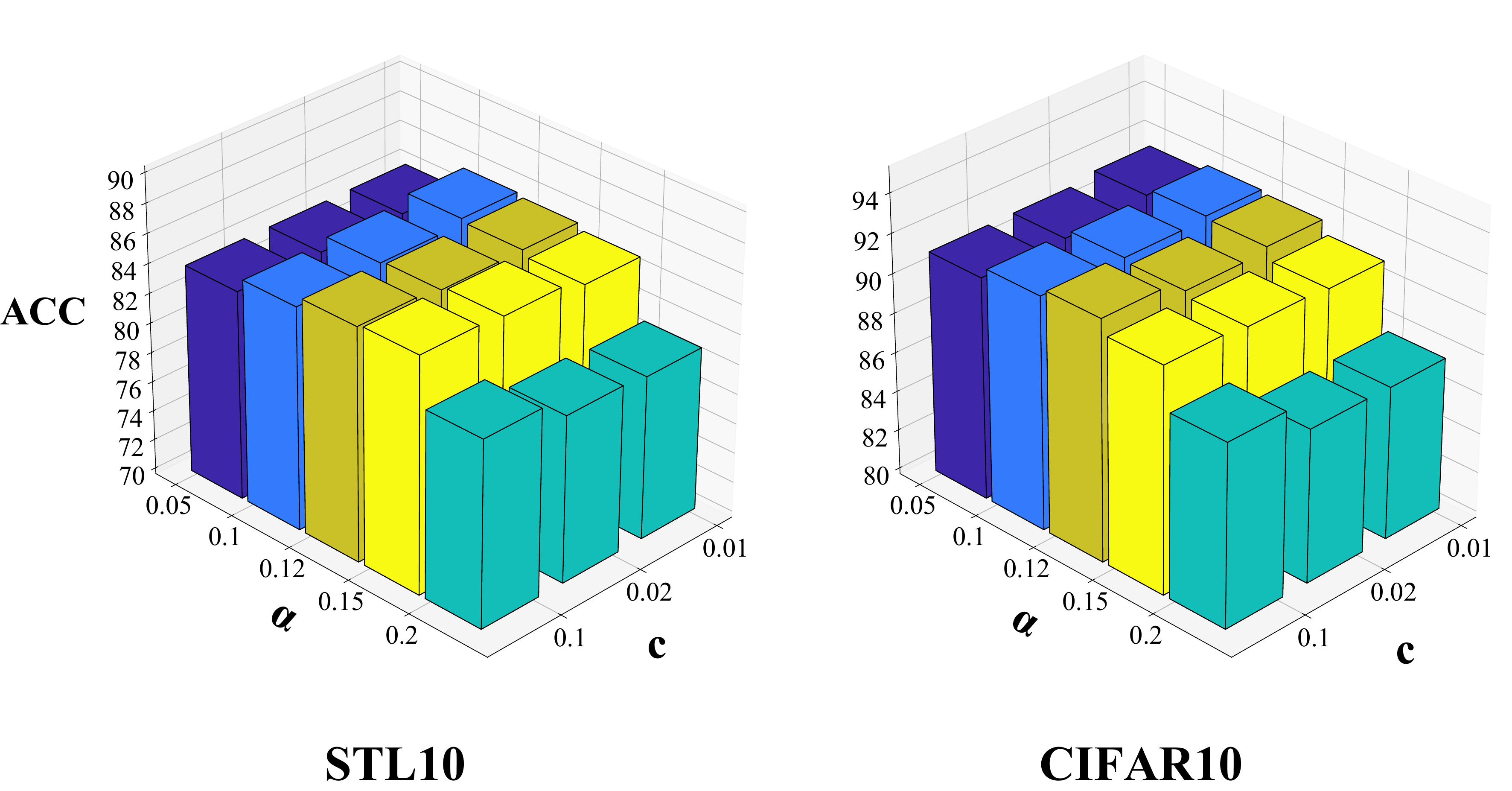}
 \label{fig:alphab}
}
\caption{\textbf{Classification accuracy with (a) different unlabeled sample size $M^u$ and $\alpha$,  (b) different $\alpha$ and $c$.} Embeddings are trained using SimCLR + \ourmethod and evaluated on CIFAR10 and STL10. For (a), we trained with different batch size (which gives different unlabeled sample size in each step) and $\alpha$, keep $c$ = 0.01. For (b), we trained with different $\alpha$ and $c$.}
\end{figure}

\subsubsection{Results for Baseline CMC ($N$=$4096$)}

Table~\ref{image acc} shows the Acc of the four methods on CIFAR10, CIFAR100 and STL10, using CMC as the baseline method. 
We observe that, (1) with a larger $N$, \ourmethod method still outperforms all the competitors as well as the baseline CMC, which verifies the generalization of \ourmethod on different sizes of negative samples $N$. (2) \emph{Hard} and \ourmethod achieve better performance than the random negative sampling method (CMC) as well as the biased negative sampling method (\emph{Ring}), which indicates the benefit of treating ``negative'' samples as unlabeled samples. (3) Compared with SimCLR and SimCLR+, CMC and CMC+ achieve worse performance, potentially due to the fact that the embedding of samples in the memory bank suffers from delayed updating that may cause noisy representations during training. (4) CMC + \ourmethod achieves 2.05\%, 4.19\% and 3.31\% improvements over the baseline CMC on CIFAR10, CIFAR100 and STL10 respectively. In all these cases CMC + \ourmethod also outperforms \emph{Hard} and \emph{Ring} with 0.62\% and 2.17\% improvements respectively.


\subsubsection{Results for Baseline Moco ($N$=$4096$)}
Table ~\ref{image acc} presents the Acc of the four methods on CIFAR10, CIFAR100 and STL10, using MoCo as the baseline method.
We observe that, (1) \ourmethod achieves the highest accuracy on all datasets and outperforms the baseline method by a substantial margin, which indicates the generalization ability of \ourmethod. (2) MoCo surpasses CMC and SimCLR on CIFAR100 and STL10, which shows the advantage of maintaining a dynamic queue of data samples. The queue removes the restriction of batch size and also discards outdated samples. (3) MoCo+\ourmethod outperforms the baseline MoCo by 5.36\%, 4.83\% and 4.55\% on CIFAR10, CIFAR100, and STL10, respectively. Compared with \emph{Hard} and \emph{Ring}, MoCo+\ourmethod also shows considerable improvements of 1.72\% and 3.24\%, respectively. (4) From these results,  we find that \ourmethod boosts all the three baseline models clearly and consistently, which verifies the efficacy of our proposed \ourmethod method regardless of the used backbone.




\subsection{Graph Representation Learning}

We compare \ourmethod with the competitors in learning graph representations via contrastive learning. We choose the state-of-the-art method InfoGraph~\cite{sun2019infograph} as the baseline and evaluate the quality of representation in graph-level classification.
We conduct all these experiments with 10 different random seeds by fine-tuning an SVM readout function and use the mean of the 10 runs as the final result. The detailed experimental settings are listed in  Appendix B and the results are summarized in Table~\ref{graph acc}. We observe that

(1) \ourmethod significantly outperforms the competitors and the baseline infoGraph, which indicates that not all the unlabeled data is active in graph datasets. (2) \ourmethod can outperform the baseline and competitors with different choices of $c$ and $\alpha$, which demonstrates the robustness of \ourmethod. 
(3) We implement two choices of $\alpha$, and the best result on 
each data shows the accuracy improvements, with 2.27\%, 5.88\%, 3.46\%, and 5.98\%.

\subsection{Ablations Studies}

We perform ablation studies on our \ourmethod framework by considering different \emph{label frequency ratio} $c$, \emph{positive prior} $\alpha$ and \emph{negative sample size} $N$ in Equation~\ref{obj:appcl} to explore their relative importance on STL10 and CIFAR10, using SimCLR as base model. 



\textbf{The effect of unlabelled sample size $M^u$.}
The unlabeled sample size $M^u$ in \ourmethod's objective Equation~\ref{obj:appcl} is for estimating the expectation of unlabeled samples. To study the effect of $M^u$, we conduct the experiments on STL10 and CIFAR10 with $M^u$ in $\{62, 126, 254, 510\}$ and illustrate results in Figure~\ref{fig:negab}. It shows that increasing $M^u$ leads to larger Acc in both datasets; the reason would be that larger $M^u$ can improve the estimation of expectation of negative distribution. This phenomenon is also observed in the competitors with large $N$ for negative samples. Therefore, we keep $N=M^u$ in all the experiments conducted in previous sections to ensure a fair comparison. 

\textbf{The effect of positive prior $\alpha$.}
$\alpha$ in \ourmethod's objective Equation~\ref{obj:appcl} represents the positive prior in unlabeled data. A good choice of $\alpha$ can result in an accurate correction in estimating the negative distributions. To study the effect of the positive prior $\alpha$ for \ourmethod, we conduct the experiments with $\alpha$ in $\{0.05, 0.1, 0.12, 0.15, 0.20\}$ and displayed the results in Figure~\ref{fig:alphab}. We observe that (1) when $\alpha$ increases from 0.05 to 0.15, the Acc value achieved by \ourmethod is consistently increased. Also, \ourmethod with different negative sample size achieves the maximum Acc in $\alpha = 0.15$ and $c=0.1$ on STL10 and obtain best performance with $\alpha = 0.12$ and $c=0.1$ on  CIFAR10. It suggests that STL10 may have a larger positive sample ratio in the unlabeled data. (2) We can observe that the best choice of $\alpha$ is between 0.1 and 0.15 on both datasets. This optimal choice of $\alpha$ is quite reasonable as both CIFAR10 and STL10 have 10 labels which would result in at least 10\% positive samples in the training dataset. In future work, we will focus on learning the prior $\alpha$.

\textbf{The effect of label frequency $c$. }
The parameter $c$ in Equation~\ref{obj:appcl} indicates the label frequency. In order to study the effect of $c$ for \ourmethod, we conduct experiments with $c$ in $\{0.01, 0.02, 0.1\}$ and the results are illustrated in Figure~\ref{fig:alphab}. Based on the results: (1) \ourmethod is not very sensitive to the choice of $c$ and the best performance with different $c$ is quite similar. For example, on STL10, the variance of the best performance for $c$ in $\{0.01, 0.02, 0.1\}$ is 0.0108\% (2) With different $c$, the best performance of \ourmethod occurs on different $\alpha$. This result suggests limited interactions between these two hyper-parameters. It is interesting to ask what might be the most optimal way for choosing those two hyper-parameters and we will leave this problem to future work.

\subsubsection{Visualization of Learned Embeddings }
To intuitively demonstrate the efficacy of \ourmethod, we visualize the learned embeddings on STL10 using UMAP~\cite{mcinnes2018umap}. We display the results in Figure~\ref{fig:emdvis}, where the color of each point represents its true label. From Figure~\ref{fig:emdvis}, we observe that \ourmethod maps observations of the same class together more frequently, which enables the downstream classifier to easily determine the class of each observations. Embeddings trained with \ourmethod also show clearer class boundaries than baseline methods.

\section{Conclusion}\label{conclusion}
In this work, inspired by Positive-Unlabeled Learning, we take the practical assumption that only the positive and unlabeled samples are available for training and the distribution of unlabeled data differs from the real distribution and propose \ourmethod, a new contrastive learning method. Both our theoretical analysis and our extensive experiments shown that \ourmethod is capable of correcting the bias introduced by the standard practice of negative sampling. \ourmethod is compatible with any algorithm that optimizes the standard contrastive loss and consistently improves the state-of-the-art baselines in various benchmarks. 
One major limitation of \ourmethod lies in the choices of two key hyperparameters: the prior $\alpha$ and the label frequency $c$. Currently, both hyperparameters are set using a grid-search method, which can be costly for large-scale datasets. In the future, we plan to design a new PU learning model which can automatically estimate $\alpha$ and $c$, especially for large datasets. Interesting directions for future work also include introducing learned correction for positive sample distribution along with the learned correction for negative sample distribution.

\bibliography{aaai24}

\end{document}